%% file: robocom_main.tex
\documentclass[conference,10pt]{IEEEtran}
\IEEEoverridecommandlockouts
\usepackage{cite}
\usepackage{amsmath,amssymb,amsfonts}
\usepackage{algorithmic}
\usepackage{graphicx}
\usepackage{textcomp}
\usepackage{xcolor}
\usepackage{enumitem}
\def\BibTeX{{\rm B\kern-.05em{\sc i\kern-.025em b}\kern-.08em
    T\kern-.1667em\lower.7ex\hbox{E}\kern-.125emX}}
\begin{document}

\title{Concurrent Transmission \\for Multi-Robot Coordination}


\author{\IEEEauthorblockN{Sourabha Bharadwaj\IEEEauthorrefmark{1},
Karunakar Gonabattula\IEEEauthorrefmark{1},
Sudipta Saha\IEEEauthorrefmark{1},\\
Chayan Sarkar\IEEEauthorrefmark{2}, and
Rekha Raja\IEEEauthorrefmark{3}}
\IEEEauthorblockA{\IEEEauthorrefmark{1} School of Electrical Sciences, Indian Institute of Technology Bhubaneswar, India}
\IEEEauthorblockA{\IEEEauthorrefmark{2}Robotics \& Autonomous Systems, TCS Research, India}
\IEEEauthorblockA{\IEEEauthorrefmark{3} Wageningen University and Research, The Netherlands}
}

\maketitle

\input{0-abstract.tex}


\input{1-introduction.tex}

\input{2-related}
\input{3-design.tex}

\input{5-expsetup.tex}
\input{6-evaluation.tex}
\input{8-conclusion.tex}

\bibliographystyle{IEEEtran}
\bibliography{robocom_main}

\end{document}

%% file: 0-abstract.tex
\begin{abstract}
An efficient communication mechanism forms the backbone for any multi-robot system to achieve fruitful collaboration and coordination. Limitation in the existing asynchronous transmission based strategies in fast dissemination and aggregation compels the designers to prune down such requirements as much as possible. This also restricts the possible application areas of mobile multi-robot systems. In this work, we introduce concurrent transmission based strategy as an alternative. Despite the commonly found difficulties in concurrent transmission such as microsecond level time synchronization, hardware heterogeneity, etc., we demonstrate how it can be exploited for multi-robot systems. We propose a split architecture where the two major activities - \textit{communication} and \textit{computation} are carried out independently and coordinate through periodic interactions. The proposed split architecture is applied on a custom build full networked control system consisting of five two-wheel differential drive mobile robots having heterogeneous architecture. We use the proposed design in a leader-follower setting for coordinated dynamic speed variation as well as the independent formation of various shapes. Experiments show a centimeter-level spatial and millisecond-level temporal accuracy while spending very low radio duty-cycling over multi-hop communication under a wide testing area.
\end{abstract}

%% file: 1-introduction.tex
\section{Introduction}
\label{sec:introduction}

In low-power IoT/Edge devices, RF communication draws a significant amount of power from the battery in comparison to the computation and other sensing operations~\cite{la2019strategies}. Although the energy consumption due to a single packet transmission is only a fraction of the energy consumption due to other operations of a mobile robot, the overall energy consumption of the communication module becomes significant with frequent and time-constrained communication. It is well-known that asynchronous transmission-based solutions for network-wide data dissemination \cite{guo2013opportunistic} wastes a lot of battery power to counter the arbitrary collisions among the uncoordinated overlapping transmissions from various nodes. They take several seconds if not minutes to disseminate a piece of information to a few communicating devices. 
On the other hand, concurrent transmission-based network-wide data dissemination to 100s of nodes requires only a fraction of a second with high reliability and lower energy consumption~\cite{ferrari2011efficient, sarkar2019fleet}. However, despite its immense capabilities, to the best of our knowledge, concurrent transmission-based strategies, due to certain specific limitations, could not be used in a general-purpose multi-robot setting so far. 

In this work, our primary objective is to achieve faster and efficient multi-robot communication by exploiting the benefits of concurrent transmission. However, the major impediment is that the concurrent transmission-based algorithms are not easily portable to any hardware. Additionally, the components of the core robotics parts are fundamentally different and independent from the communication module. This causes additional computation time and cannot uphold the microsecond level synchronization among the communicating devices, which is the absolute necessity for concurrent transmission to succeed. To bring an easy and elegant synergy between both the issues, in this work, we propose a physical layer separation of the \textit{computation and control system} part from the special \textit{communication module} supporting  concurrent transmission. These two modules are termed as \textit{Computation and Control Unit (CCU)} and \textit{Networking Unit (NU)}. To demonstrate the effectiveness of the proposed setting, we use a well-known concurrent transmission-based one-to-all dissemination protocol \textit{Glossy}~\cite{ferrari2011efficient} for fast and energy-efficient coordination between a single leader and multiple follower robots under multi-hop connectivity.

In summary, the contributions of this work are as follows. 

\begin{itemize}[leftmargin=*]
    \item We introduce concurrent transmission as an efficient and fast mechanism for communication among multiple robots. We demonstrate an easy and flexible way to achieve the same.
    
    \item We propose a separation between CCU and NU and their loose coupling to solve the core issues in porting concurrent transmission.

    \item We provide a detailed study of our system for a robotic control system using a multi-hop, \textit{single-leader-multiple-follower} setting. The implementation in hardware platforms demonstrates the feasibility of fairly accurate real-time coordination between the leader and the follower with a millisecond-level time synchronization.
\end{itemize}

%% file: 2-related.tex
\section{Related work}
\label{sec:relatedwork}

Multi-robot communication and control-system design have been addressed by many works so far in different ways such as leader-follower strategy \cite{li2016affection, lin2020adaptive}, virtual structure approach \cite{nazari2016decentralized, wu2020decentralized} and behavior-based method \cite{balch1998behavior, yu2019formation} etc. However thse works are either based on the theoretical studies \cite{pereira2003formation, chen2011leader, loria2015leader, panagou2014cooperative} or implemented in simulation platforms \cite{chehardoli2018third, maghenem2018robust}. Very few works are done in practical setups \cite{wurman2008coordinating, widyotriatmo2017implementation}. But these works lack the required communication latency to support underlying control strategies.

Under traditional asynchronous transmission-based MAC strategies, the nodes usually compete with each other to gain access to the medium. They waste a lot of bandwidth and energy in the nodes to mitigate the collisions among the packets from different nodes. In contrast, under the \textit{concurrent transmission} paradigm, the nodes cooperate in such a way that the RF packets emanating from different devices are transmitted exactly at the same time resulting in either \textit{capture effect} or \textit{constructive interference} in the physical layer of the network stack~\cite{rao2016murphy}. 

Many fundamental communication problems are solved very efficiently using the concurrent transmission~\cite{jacob2019synchronous}, e.g., one-to-all dissemination is achieved by Glossy \cite{ferrari2011efficient}, many-to-many data sharing and data aggregation is done by Chaos \cite{kaddoum2013design}, all-to-all and many-to-one communication is achieved by LWB \cite{sarkar2016lwb}, MiniCast \cite{minicast} etc. These protocols can achieve reliability of 99.99\% with ultra-low radio power consumption and millisecond-level latency, even when 100s of nodes communicate over a multi-hop setup. Owing to such paramount capabilities, we propose to use concurrent transmission-based strategies for efficient multi-robot communication and collaboration. 

%% file: 3-design.tex
\section{Design}\label{sec:design}

\begin{figure}[t!]
\begin{center}
\includegraphics[width=\linewidth]{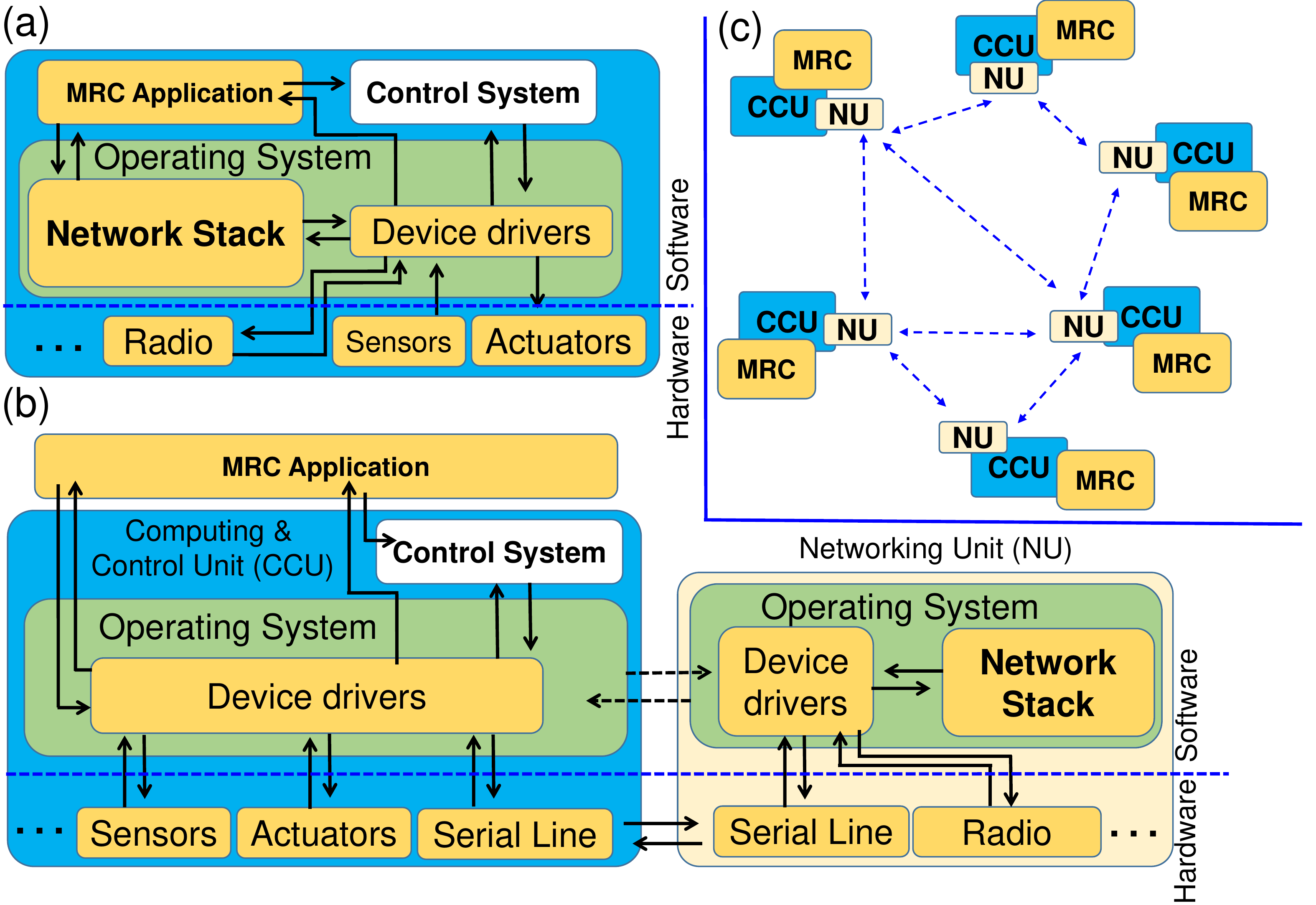}
\end{center}
\caption{Part (a) shows the general modules of an independent networked robotic system. Part (b) shows the separation of the \textit{Network Unit} (NU) and \textit{Computing \& Control Unit} (CCU). Part (c) shows a network among multiple independent robotic systems. MRC stands for \textit{Multi-Robot Coordination}.} 
\label{fig:separation}
\vspace{-0.5cm}
\end{figure}

In this section, we detail the basic design step for a hardware level separation of the computation/control and communication, and the serial line protocol to support loose coupling between these two modules to achieve flexibility.

\subsection{Separation of computation and communication} 
Concurrent transmission based strategies fundamentally use special physical layer phenomena such as constructive interference and capture effect. They were shown to be very much successful in a few specific devices. One of them is \textit{TelosB}, mainly composed of an MSP430 micro-controller along with a CC2420 radio chip compatible with 802.15.4 standards. Some new devices are also available following the same architecture. A large number of prevalent works have been tested on this architecture, and all reported a very consistent behavior~\cite{almalkawi2010wireless}. However, although a simple micro-controller structure used in these devices is sufficient to implementation of ST very successfully, they are neither suitable to carry out complex computations necessary for realizing the control systems nor the CPU and memory intensive operations. In ST, the required software architecture also is quite different and hence is not that suitable for a general-purpose computation platform. 

Thus, in order to gain the benefit of both sophisticated computing as well as reliable, fast, and energy-efficient communication framework, in this work we propose, a hardware (and software) level separation of the two main units - \textit{Computation and Control Unit} (CCU) and \textit{Communication Unit} (NU). The system is supposed to work through a loose coupling between the two units discussed later. Fig.~\ref{fig:separation} shows a communication network under this split mode where the NUs would only communicate with each other. In general, the two units CCU and NU represent two different independent domains, and hence, this split architecture would naturally allow their independent development and easy integration between them.

In this work, we mainly use the \textit{TelosB} devices for NUs and various off-the-shelf standard devices such as \textit{Raspberry Pi}, \textit{Beaglebone}, etc., as CCUs. Ideally, any other suitable device, such as \textit{Arduino}, \textit{Jetson}, can also act as CCU, for different types of robots. We also test our base strategy with the role of CCUs being played by general-purpose desktop computers and laptops. In our work, we use Ubuntu operating system in the CCUs and the Contiki OS in the NUs. 

\begin{figure}[t!]
\begin{center}
\includegraphics[width=\linewidth]{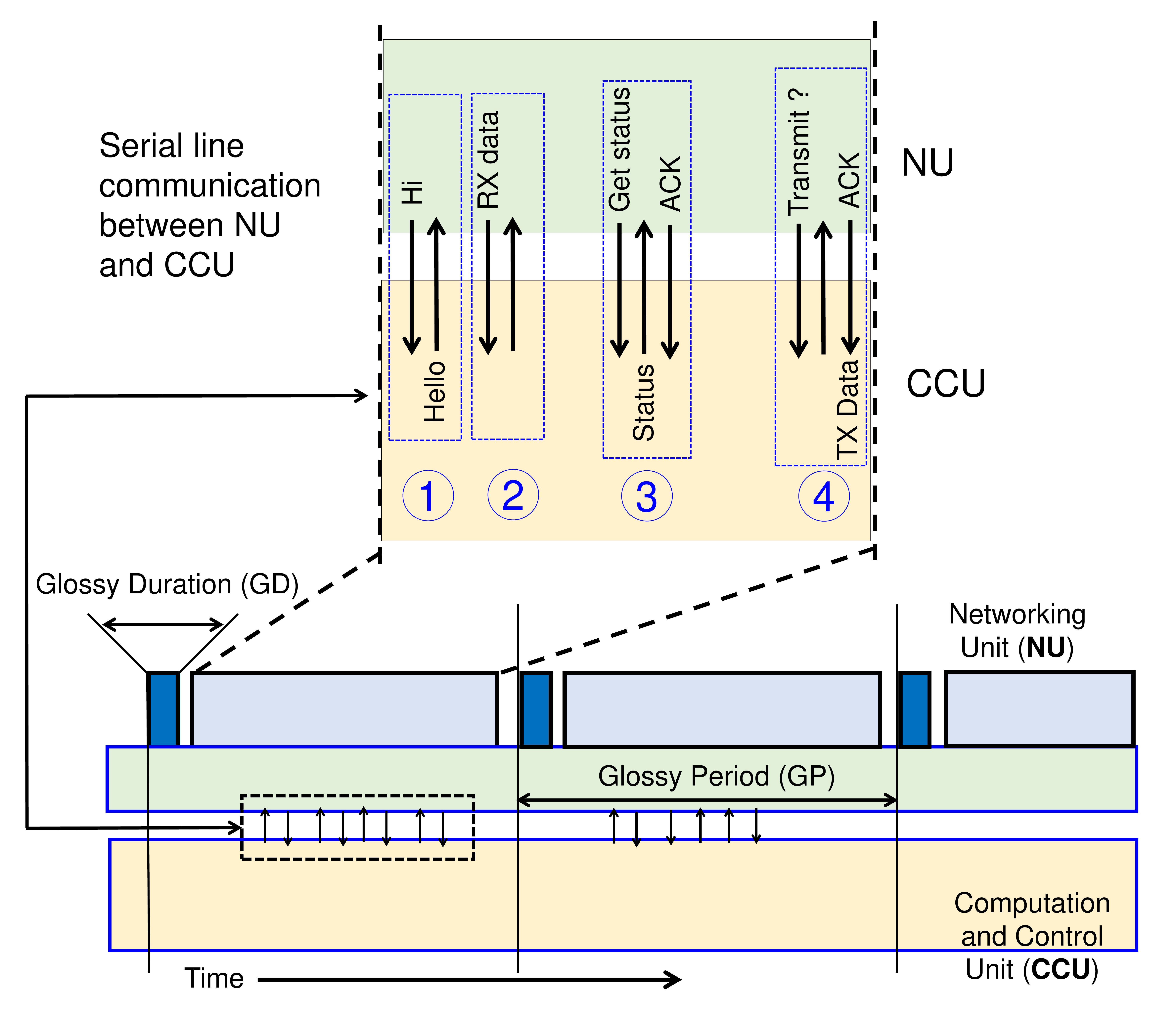}
\end{center}
\caption{Four different phases, (marked by circled numbers) in the communication between the NUs and the CCUs using the the serial line protocol.} 
\label{fig:slp}
\vspace{-0.5cm}
\end{figure}

\subsection{Serial line protocol (SLP)}
In our split architecture, the NUs talk to each other through standard protocols using concurrent transmission. For the communication between a CCU and a NU, in this work, we use a loosely coupled serial line communication.

Under any concurrent transmission based strategy, communication happens in a periodic fashion. Time is divided into slots of equal length (say $p$ time units). A small part of this slot ($d$ time units) is used for time synchronization as well as NU-NU data communication. All the devices set their radios ON during this part. In each slot, a small part is dedicated to communication among the devices. The remaining time in a slot, the radio remains OFF. In the NUs, we exploit this part for the sake of NU-CCU serial line communication. However, note that it's hard for a CCU to keep track of the fine-grained division of the time slots implemented by the NUs. Therefore, in this setup, we assign the responsibility of initiating the NU-CCU communication to NUs. The CCU-NU interaction is achieved in a query-reply form where NU initiates the queries and CCU responds with the replies. 

We implement the CCU-NU interactions in four steps. First, the exchange of formal `Hi' and `Hello' messages are done to ensure availability. Once found to be in good status the NU transfers the data to CCU in the second step. This data is read by the NU in the NU-NU communication, which is in turn prepared by the source CCU at least $d$ unit of time early. Therefore, the first two steps of NU-CCU interactions are executed immediately after the NU-NU communication part is over to ensure the data received by the CCUs to be as fresh as possible.
The third step is performed approximately in the middle of the period. This part is explicitly used for a synchronized recording of data and uses the same for offline analysis for calculation of the metrics. Usually, critical and state-related information is recorded in this part. Step 4 is executed as much as possible towards the end of the period. This part is used to record the necessary CCU state information as decided by the MRC algorithm and hand over the same to the NU so that in the next NU-NU communication the same can be shared with other NUs. All these four steps are pictorially described in Fig.~\ref{fig:slp}.


%% file: 5-expsetup.tex
\section{Experimental setup and metrics}\label{sec:expsetup}
Several experiments have been carried out using the proposed framework for multi-robot communication for five different robots. We explore two possible settings (a) \textit{central-controller based}, (b) \textit{leader-follower based}.

Under the central-controller based setting, a separate static computer connected with an NU is designated as a central-controller which sends the command as and when necessary. The commands get disseminated through a multi-hop communication network formed by the NUs and ultimately reach the CCUs which finally execute the commands. 

Under the leader-follower setting, one of the devices works as a leader while the others act as the follower. We use differential wheel drive cars where the main actuators are the motors to govern the wheels. The PID loop of the leader begins with the reading of the current status of the actuators. It calculates the difference and the necessary voltage value to be set to achieve the reference speed based on the well-known PID controlling rules. The leader applies the necessary changes and follows the loop to independently decide what is to be done in the next iterations onward. The MRC application running in the follower decides what is the next reference value that it should set for its actuators based on the leader's current speed.

\begin{figure}[t!]
\begin{center}
\includegraphics[width=0.5\textwidth]{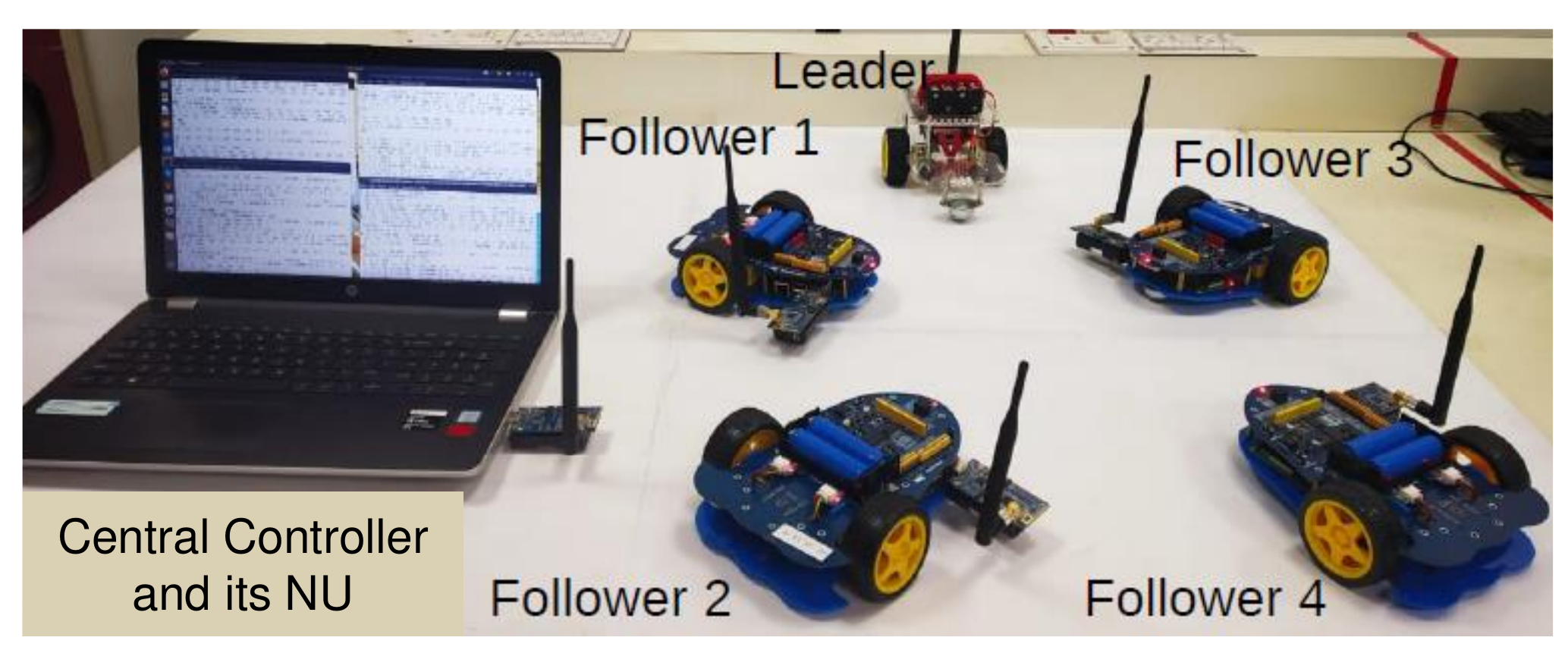}
\end{center}
\caption{
Our experimental setup with 5 devices and a laptop. The laptop works as the central controller and in general it is used for the collection of log information for analysis purpose.} 
\label{fig:our_exp_setup}
\vspace{-0.5cm}
\end{figure}

For our experiments, we use one \textit{GoPiGo} car and four \textit{AlphaBot} cars (Fig.~\ref{fig:our_exp_setup}). To support our proposed split-architecture (discussed in Section \ref{sec:design}A), we mount the bots with a Raspberry Pi unit and TelosB unit to carry out the computations (CCU) and communication (NU), repectively.

\noindent\textit{\textbf{A. Parameters}}: The major parameters are as follows. 

\textit{Glossy Duration (GD) and Glossy Period (GP)}: All the nodes (NUs) participating in the process carry out the radio-packet transmission and reception as per the Glossy protocol for a stipulated maximum time, termed as Glossy Duration (GD) (Fig.~\ref{fig:slp}). The radios in all the devices can stay ON for a maximum of GD period. After that, the nodes turn off their radio and resumes after a period. This repeated resuming time is termed as Glossy Period (GP).

\textit{PID Period (PP)}: The control system in robots uses a loop (PID-loop) for periodic calculation of errors and adjustments of the speeds through appropriate variation of the voltage/current inputs. The PID-loop is characterized by a parameter termed as PID period (PP).

\noindent\textit{\textbf{B. Metrics}}:
In most of our experiments, we set some goals for the devices, e.g., target speed for the devices, and then observe in the experiments how well the devices achieve the goal. To quantify the performance, we calculate the difference between the target and the achieved goals and subsequently measure the percentage of the deviation concerning the target. We calculate the following two average error percentages for the whole duration of the experiment.

\textit{PID-err}: PID-err is the percentage of error of the actual speed achieved by the device and the target speed. It depends on the PP of the PID-loop running in each of the devices.

\textit{TRX-err}: Our experiments involve the communication of instructions from a leader node or a central controller. Under a multi-hop setting, we transmit the measured speed value through the Glossy protocol itself and compare the same with the achieved speed and calculate the average percentage error in the same way. We call this TRX-err. Thus, it depends on both the Glossy loop and the PID-loop, i.e., the GP and PP  values, respectively. 


%% file: 6-evaluation.tex
\section{Evaluation}\label{sec:evaluation}

We perform three sets of experiments based on the proposed design. First, we test the extent of time synchronization achieved among the devices. Next, we find the optimal PP value for further experiments. After that, we try to find the best GP-PP combination by studying the interaction between the control and communication sub-systems. 

\subsection{Time synchronization among the CCUs}
Concurrent transmission based one-to-all dissemination protocol that we use in our work provides a micro-second level time synchronization among the NUs. Our split architecture makes it possible to extend this precise and accurate time synchronization in all the CCUs under consideration. Although the target of the work is not to focus on perfect time synchronization, it's interesting to check how well the loose serial line-based coupling can enable the CCUs to make use of the said synchronization of time.

In a general setting, the accurate measurement of a time synchronization strategy is usually done in two steps. First, a specific output PIN from the micro-controller of the devices being synchronized (here the CCUs), is set to go high when the probe for the reference time is received. Next, these indicator PINs are connected to a single oscilloscope to measure the difference in the times when the PINs go high in different devices through several iterations. The authors of \cite{ferrari2011efficient} demonstrate a micro-second level accurate time synchronization in the protocol Glossy using this strategy. However, under our proposed scheme, unlike the Glossy setup, the CCUs and NUs are supposed to be very loosely coupled and operate independently through simple serial line communication without any use of interrupt from none of the sides. Thus, the target here is to understand how well time synchronization can be achieved amidst such support for heterogeneity and flexibility.

We use the standard network time synchronization protocol NTP \cite{theNTP} to evaluate the synchronization accuracy achieved. NTP takes the help of the internet and connects to the remote server machines to fetch time synchronization information to the nodes connected as clients. The accuracy level of NTP is known to be of the order of milliseconds (not microsecond level). In this experiment, we use five general-purpose computers, including desktop (Intel core-i7 processors, with 8GB RAM), high-end workstations (Intel Xeon multi-core processors and 64GB RAM), three laptops, as well as five different Raspberry PIs and five Beaglebone devices. In all these devices we install a compatible Ubuntu operating system and configure NTP. We consider all these machines as heterogeneous CCU platforms. The NU devices were connected with a serial line with each of these CCUs. These CCUs (along with the connected NUs) were placed at various locations on the same floor of a building covering an area of about 236 ft X 84 ft. The NUs formed a three-hop network under this setting. The NUs run the standard Glossy protocol for flooding along with time synchronization and communicate with the attached CCUs through the serial line based presented in Section \ref{sec:design}. In the first step of every round of CCU-NU interaction, i.e., per every Glossy period, when the NUs send a \textit{Hi} message, the respective CCU records the local timestamp (synced by NTPs) of that message. These log messages are collected from all the devices and differences are calculated in these time stamps offline. The process is repeated for different Glossy periods. 

\begin{figure}[t!]
\begin{center}
\includegraphics[width=\linewidth]{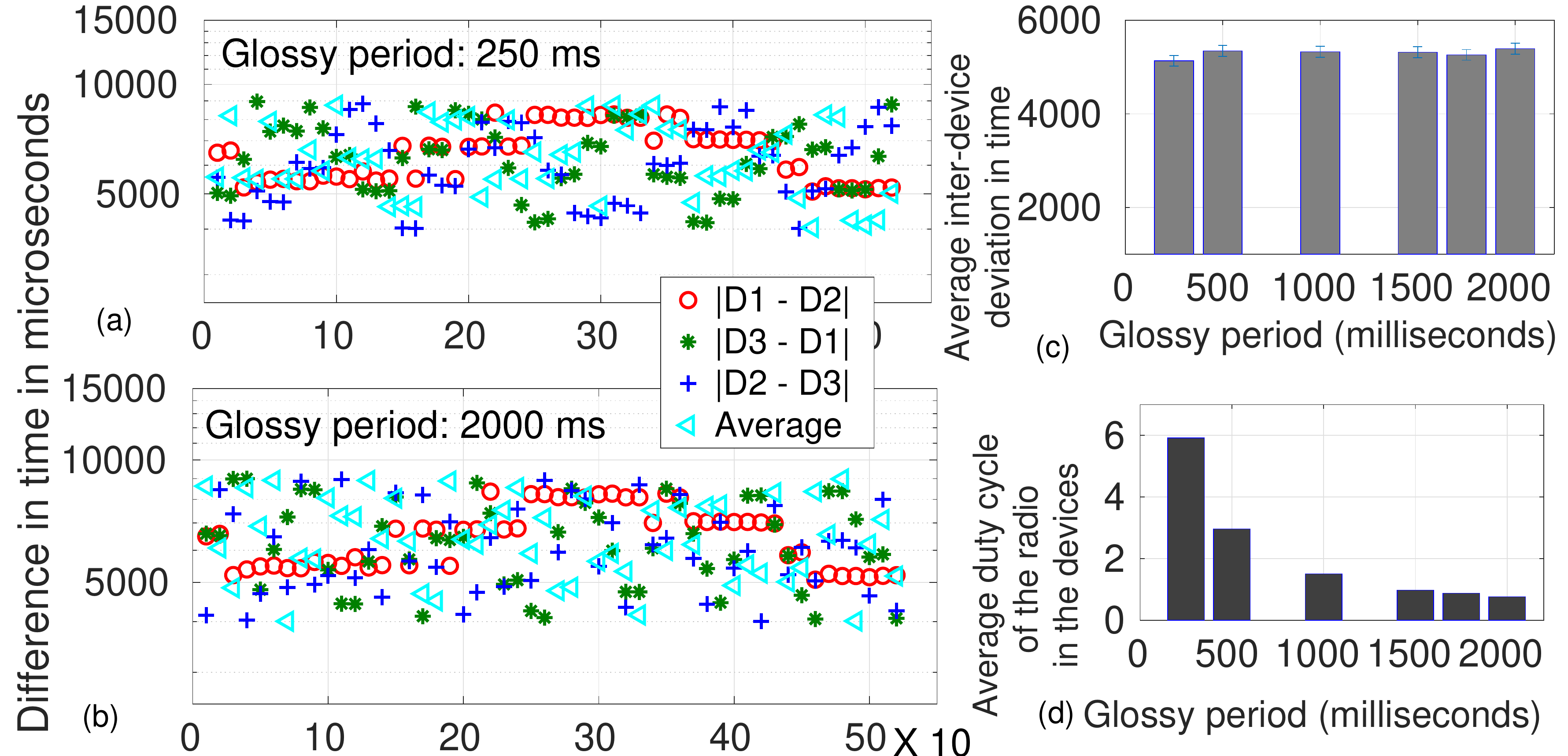}
\end{center}
\caption{
Part (a) shows sample time-deviation between device 1 and device 2 for a GP of 2 sec. Part (b) shows sample differences of time between device 1 and device 2 for GP 250 ms ($Di$ in the legend indicates the time in Device $i$). Average time differences among all the devices along with standard deviation as error bars are shown in part (c). Part (d) shows the average radio-duty cycle for different GPs.
} 
\label{fig:sync}
\vspace{-0.5cm}
\end{figure}

Fig.~\ref{fig:sync} plots some of the sample average time differences for different GPs. It can be observed despite loose coupling between NUs and CCUs, as well as considering NTP as the ground truth (which itself has a deviation in millisecond level), the synchronization error in our proposed schemes does not go beyond max 12 ms with an average deviation of 5 ms. We also observe that the achieved accuracy is almost the same irrespective of any value of GP. The radio duty-cycling for various GP settings in the experiment is also shown in the same (Fig.~ \ref{fig:sync}(d)). We calculate the radio duty-cycling as the ratio of the values of the GD to the GP. While keeping the GD constant and increasing the GP, the sharp decrease in the radio duty-cycling is quite visible in the figure. 

Note that, in a decentralized and infrastructure-less scenario, without any help of the Internet, even millisecond-level time synchronization is extremely important. The synchronization can be improved further with the help of a careful setting of interrupts. But the target of the current experiment is to understand what is the baseline without using any such sophisticated strategy.

\subsection{Independent study of the control system}
We study how PP affects the maximum speed targeted by a device in our setup. The PP of a device is varied locally. For each PP and target speed combination, we execute 10 runs each having a duration of 2 minutes. In each case, we compute the difference between the achieved and the target speed and calculate the average of these errors. In a nutshell, we find that the average error is high both when a high PP is set for low target speed as well as a low PP is set for high target speed. The relation between the PP and the speed that can be achieved under that PP is almost linear. It can be concluded that to achieve higher speed in a smooth and error-free way, one should choose larger PP values and vice versa.

\begin{figure}[t!]
\begin{center}
\includegraphics[width=\linewidth]{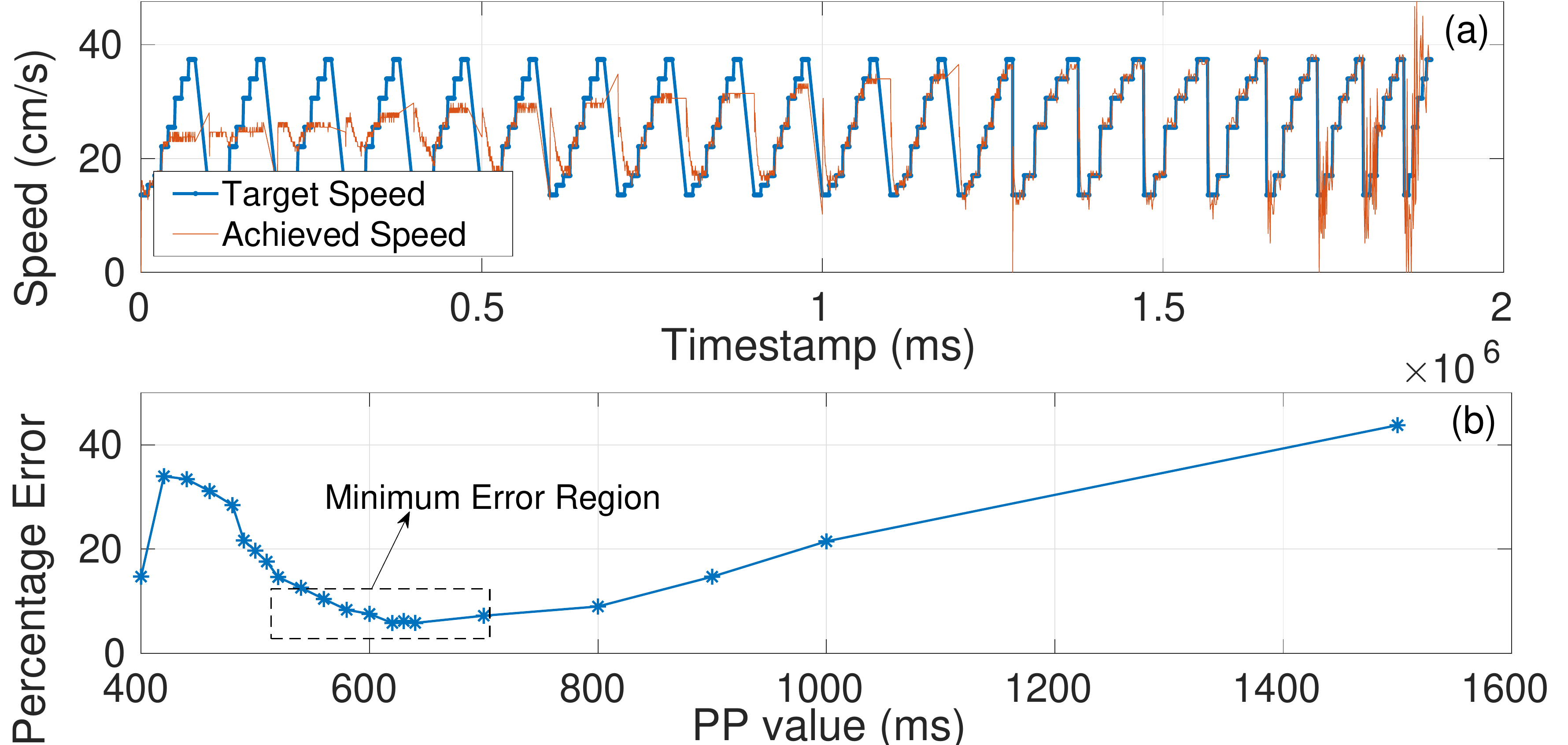}
\end{center}
\caption{Part (a) represents device performance for different range of PP settings. Each triangular segment in the plot represents an experiment run for a certain PP value. Part (b) represents the corresponding average percentage errors of the device for respective PP values in part (a).} \label{fig:local_pp_change_ts_and_err}
\vspace{-0.5cm}
\end{figure}

However, dynamically changing the PP is hard. Therefore, in the next set of experiments, we try to find a single PP that can produce the best result under a wide range of target speeds. We vary the PP of the robots from 400 ms to 1500 ms, and for each PP, the target speed is varied from 14 cm/s to 34 cm/s. We measure the error in each of the cases as the absolute difference between the target speed and achieved speed with time. The average error values are plotted in Fig.~\ref{fig:local_pp_change_ts_and_err}.

We observe that the average error decreases until a particular range of PP. But after that, it shoots up, forming a parabolic curve. With this, we could conclude that setting the PP in a range 500 ms - 700 ms provides the least error under the said range of speeds. In the next set of experiments, we use 600 ms for our PP.


\subsection{Interaction between control and communication}

All ST based communication strategies depend on periodic time synchronization among the NUs. This periodic activity we term as \textit{sync-loop}. GP and PP act as the period of the sync-loop and the PID loop. 
We carry out a series of experiments through which we try to understand the dynamics of the interaction between sync-loop and PID-loop as detailed below.

\begin{figure}[t!]
\begin{center}
\includegraphics[width=\linewidth]{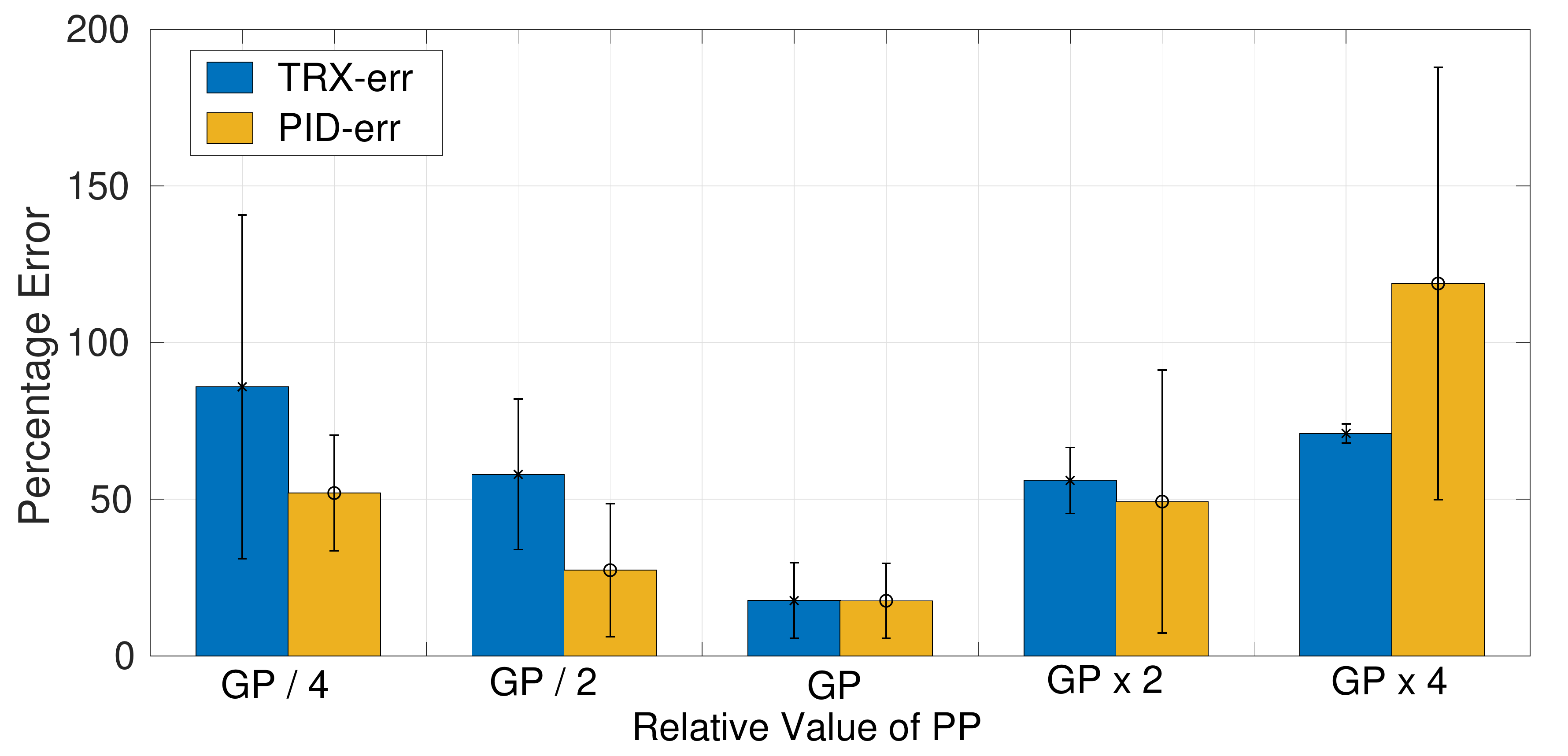}
\end{center}
\caption{Effect of various GP-PP combination under central-controller based experiments.} 
\label{fig:cmd_gp_pp_scale_err}
\vspace{-0.5cm}
\end{figure}

\textbf{Central-controller based experiment}: We first conduct the command based experiment (see Section \ref{sec:expsetup} for details). GP is varied from 200 ms to 1100 ms in steps of 100 ms. For each GP, the PP is varied as GP/4, GP/2, GP, 2GP, and 4GP. For each such GP-PP combination, the target speed is varied from 0 cm/s to 34 cm/s for every 20 seconds. In each case, we calculate the average error values. Fig.~\ref{fig:cmd_gp_pp_scale_err} shows PID-err and TRX-err (see Section \ref{sec:expsetup}B for details) vs PP values scaled with respect to the GP values. We show these through a barplot in Fig.~\ref{fig:cmd_gp_pp_scale_err} to highlight the importance of choosing appropriate GP and PP values. The PID-loop is quintessential for updating the speed at the hardware level. Thus, if a device needs to communicate its speed, it should be the most recent one, i.e., the one updated by the PID-loop. Therefore, for combinations of a low PP and high GP as well as high PP and low GP, does not perform as expected. Based on the results obtained from all these experiments, it can be seen that the best performance from the proposed split architecture can be obtained when \textbf{PP is the same as GP}. It can also be observed that having a very high GP value although is good in terms of duty cycling in the devices, is not good in terms of the underlying control activities. The range of GP, 500 ms - 700 ms, performs the best. Following these observations, in all rest of our experiments, we set the GP value and PP value to 600 ms. Note that, in all the results for the rest of the experiments, we present the average of the two errors, i.e., PID-err and TRX-err.


\begin{figure}[ht!]
\begin{center}
\includegraphics[width=\linewidth]{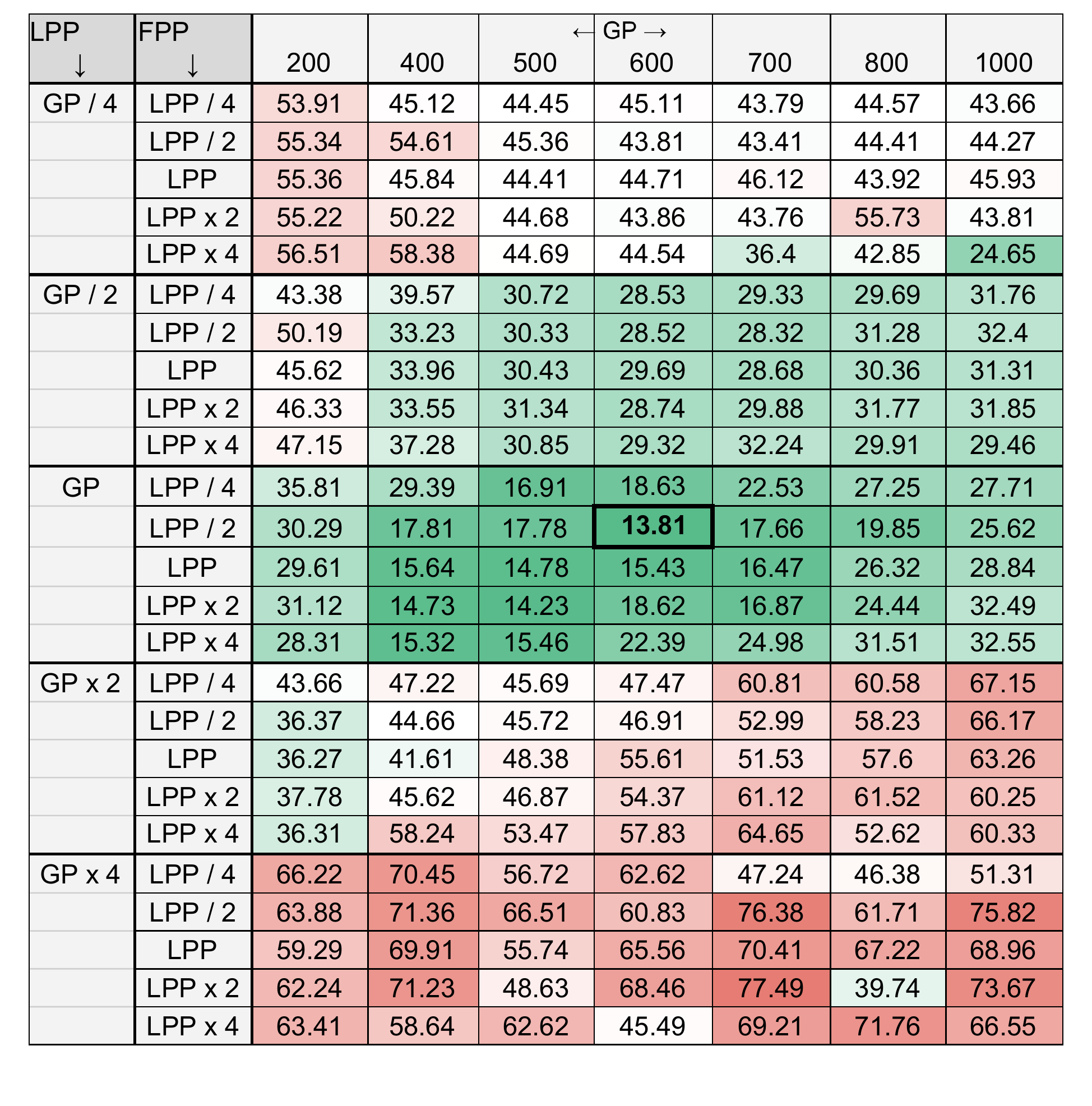}
\end{center}
\caption{Heatmap showing the error percentages of various combination of GP, LPP and FPP under leader-follower based operation. The columns represent the GP values. The rows are grouped based on various values of LPP presented in terms of GP. In each group the values of the FPPs are shown with respect to the LPP value. The best performance is observed when LPP equals GP and FPP equals LPP/2.} \label{fig:hmap_lf_gp_pp_setting}
\vspace{-0.5cm}
\end{figure}

\textbf{Leader-follower based experiment}: Next, we verify how do the observations from the previous section hold in the case of leader-follower based setup. Under this setup, apart from the GP which is the same in the leader and the followers, there are two other parameters - the leader's PP and the followers' PP, abbreviated as LPP and FPP, respectively. We carry out a series of experiments to find out the optimal combination and derive the best suitable combination for these three parameters as detailed below.

We test different combinations of the Glossy and the leader-follower PID cycles. GP is varied from 200 ms to 1000 ms. For each such GP, the LPP is varied from GP/4 to 4GP. For each LPP, we vary the FPP from LPP/4 to 4LPP. Finally, for each such GP-LPP-FPP setting, the target speed is varied over the following values 0, 14, 34, 20 (all units are in cm/s), with every speed lasting for 12 seconds. The average percentage error values computed in these experiments are presented in the Fig.~\ref{fig:hmap_lf_gp_pp_setting} in the form of a heatmap. It is clearly observable that in our setting the least error is achieved at GP = 600 ms while LPP equals to GP, i.e., 600 ms and, and FPP equals to half the LPP, i.e., 300 ms. Thus, in summary, running the leader PID-loop in synchronization with the sync-loop and running the follower PID-loop two-times faster than the leader PID-loop produces the best performance. The reason behind this observation is as follows. The followers always receive the leader's updated speed in the next Glossy loop. This implies that a constant lag exists between the leader's and the followers' speed. If we update the speed of the follower at the same rate as that of the leader, this lag would extend further. Hence, we need to run the PID loop of the follower relatively faster so that at least the effect of the lag can be minimized to a reasonable extent.

%% file: 8-conclusion.tex
\section{Conclusion}
\label{sec:conclusion}

Microsecond level time synchronization and a few specific hardware dependencies makes the energy-efficient and robust concurrent transmission based strategies hard to be applicable in a generic networked systems like multi-robot system. In this article, we propose a loosely coupled split architecture to make use of concurrent transmission for multi-robot systems even under infrastructure-less fully decentralized settings. The proposed strategy is fully implemented over off-the-shelf robotic platforms and extensively tested through various experiments under both central-controller based as well as leader-follower based settings. 

%% file: robocom_main.bbl
\begin{thebibliography}{10}
\providecommand{\url}[1]{#1}
\csname url@rmstyle\endcsname
\providecommand{\newblock}{\relax}
\providecommand{\bibinfo}[2]{#2}
\providecommand\BIBentrySTDinterwordspacing{\spaceskip=0pt\relax}
\providecommand\BIBentryALTinterwordstretchfactor{4}
\providecommand\BIBentryALTinterwordspacing{\spaceskip=\fontdimen2\font plus
\BIBentryALTinterwordstretchfactor\fontdimen3\font minus
  \fontdimen4\font\relax}
\providecommand\BIBforeignlanguage[2]{{%
\expandafter\ifx\csname l@#1\endcsname\relax
\typeout{** WARNING: IEEEtran.bst: No hyphenation pattern has been}%
\typeout{** loaded for the language `#1'. Using the pattern for}%
\typeout{** the default language instead.}%
\else
\language=\csname l@#1\endcsname
\fi
#2}}

\bibitem{la2019strategies}
R.~La~Rosa, P.~Livreri, C.~Trigona, L.~Di~Donato, and G.~Sorbello, ``Strategies
  and techniques for powering wireless sensor nodes through energy harvesting
  and wireless power transfer,'' \emph{Sensors}, vol.~19, no.~12, p. 2660,
  2019.

\bibitem{guo2013opportunistic}
S.~Guo, L.~He, Y.~Gu, B.~Jiang, and T.~He, ``Opportunistic flooding in
  low-duty-cycle wireless sensor networks with unreliable links,'' \emph{IEEE
  Transactions on Computers}, vol.~63, no.~11, pp. 2787--2802, 2013.

\bibitem{ferrari2011efficient}
F.~Ferrari, M.~Zimmerling, L.~Thiele, and O.~Saukh, ``Efficient network
  flooding and time synchronization with glossy,'' in \emph{In
  \textit{Proceedings of} ACM/IEEE IPSN}.\hskip 1em plus 0.5em minus
  0.4em\relax IEEE, 2011, pp. 73--84.

\bibitem{sarkar2019fleet}
C.~Sarkar, R.~V. Prasad, and K.~Langendoen, ``Fleet: When time-bounded
  communication meets high energy-efficiency,'' \emph{IEEE Access}, vol.~7, pp.
  77\,555--77\,568, 2019.

\bibitem{li2016affection}
F.~Li, Y.~Ding, M.~Zhou, K.~Hao, and L.~Chen, ``An affection-based dynamic
  leader selection model for formation control in multirobot systems,''
  \emph{IEEE Transactions on Systems, Man, and Cybernetics: Systems}, vol.~47,
  no.~7, pp. 1217--1228, 2016.

\bibitem{lin2020adaptive}
J.~Lin, Z.~Miao, H.~Zhong, W.~Peng, Y.~Wang, and F.~Rafael, ``Adaptive
  image-based leader-follower formation control of mobile robots with
  visibility constraints,'' \emph{IEEE Transactions on Industrial Electronics},
  2020.

\bibitem{nazari2016decentralized}
M.~Nazari, E.~A. Butcher, T.~Yucelen, and A.~K. Sanyal, ``Decentralized
  consensus control of a rigid-body spacecraft formation with communication
  delay,'' \emph{Journal of Guidance, Control, and Dynamics}, vol.~39, no.~4,
  pp. 838--851, 2016.

\bibitem{wu2020decentralized}
B.~Wu, C.~Xu, and Y.~Zhang, ``Decentralized adaptive control for attitude
  synchronization of multiple spacecraft via quantized information exchange,''
  \emph{Acta Astronautica}, 2020.

\bibitem{balch1998behavior}
T.~Balch and R.~C. Arkin, ``Behavior-based formation control for multirobot
  teams,'' \emph{IEEE transactions on robotics and automation}, vol.~14, no.~6,
  pp. 926--939, 1998.

\bibitem{yu2019formation}
H.~Yu, P.~Shi, C.-C. Lim, and D.~Wang, ``Formation control for multi-robot
  systems with collision avoidance,'' \emph{International Journal of Control},
  vol.~92, no.~10, pp. 2223--2234, 2019.

\bibitem{pereira2003formation}
G.~A. Pereira, A.~K. Das, V.~Kumar, and M.~F.~M. Campos, ``Formation control
  with configuration space constraints,'' in \emph{In In \textit{Proceedings
  of} IEEE IROS}, vol.~3.\hskip 1em plus 0.5em minus 0.4em\relax IEEE, 2003,
  pp. 2755--2760.

\bibitem{chen2011leader}
G.~Chen and F.~L. Lewis, ``Leader-following control for multiple inertial
  agents,'' \emph{International Journal of Robust and Nonlinear Control},
  vol.~21, no.~8, pp. 925--942, 2011.

\bibitem{loria2015leader}
A.~Loria, J.~Dasdemir, and N.~A. Jarquin, ``Leader--follower formation and
  tracking control of mobile robots along straight paths,'' \emph{IEEE
  transactions on control systems technology}, vol.~24, no.~2, 2015.

\bibitem{panagou2014cooperative}
D.~Panagou and V.~Kumar, ``Cooperative visibility maintenance for
  leader--follower formations in obstacle environments,'' \emph{IEEE TRO},
  vol.~30, no.~4, pp. 831--844, 2014.

\bibitem{chehardoli2018third}
H.~Chehardoli and M.~R. Homaeinezhad, ``Third-order leader-following consensus
  protocol of traffic flow formed by cooperative vehicular platoons by
  considering time delay: constant spacing strategy,'' \emph{Proceedings of the
  Institution of Mechanical Engineers, Part I: Journal of Systems and Control
  Engineering}, vol. 232, no.~3, pp. 285--298, 2018.

\bibitem{maghenem2018robust}
M.~Maghenem, A.~Loria, and E.~Panteley, ``A robust $\delta$-persistently
  exciting controller for leader-follower tracking-agreement of multiple
  vehicles,'' \emph{European Journal of Control}, vol.~40, pp. 1--12, 2018.

\bibitem{wurman2008coordinating}
P.~R. Wurman, R.~D'Andrea, and M.~Mountz, ``Coordinating hundreds of
  cooperative, autonomous vehicles in warehouses,'' \emph{AI magazine},
  vol.~29, no.~1, 2008.

\bibitem{widyotriatmo2017implementation}
A.~Widyotriatmo, E.~Joelianto, A.~Prasdianto, H.~Bahtiar, and Y.~Y. Nazaruddin,
  ``Implementation of leader-follower formation control of a team of
  nonholonomic mobile robots,'' \emph{IJCCC}, vol.~12, no.~6, 2017.

\bibitem{rao2016murphy}
V.~S. Rao, M.~Koppal, R.~V. Prasad, T.~V. Prabhakar, C.~Sarkar, and
  I.~Niemegeers, ``Murphy loves ci: Unfolding and improving constructive
  interference in wsns,'' in \emph{IEEE INFOCOM 2016-The 35th Annual IEEE
  International Conference on Computer Communications}.\hskip 1em plus 0.5em
  minus 0.4em\relax IEEE, 2016.

\bibitem{jacob2019synchronous}
R.~Jacob, J.~B{\"a}chli, R.~Da~Forno, and L.~Thiele, ``Synchronous
  transmissions made easy: Design your network stack with baloo,'' in \emph{In
  \textit{Proceedings of} EWSN}.\hskip 1em plus 0.5em minus 0.4em\relax
  Junction Publishing, 2019, pp. 106--117.

\bibitem{kaddoum2013design}
G.~Kaddoum, F.-D. Richardson, and F.~Gagnon, ``Design and analysis of a
  multi-carrier differential chaos shift keying communication system,''
  \emph{IEEE Transactions on Communications}, vol.~61, no.~8, 2013.

\bibitem{sarkar2016lwb}
C.~Sarkar, ``Lwb and fs-lwb implementation for sky platform using contiki,''
  \emph{arXiv preprint arXiv:1607.06622}, 2016.

\bibitem{minicast}
S.~Saha, O.~Landsiedel, and M.~C. Chan, ``Efficient many-to-many data sharing
  using synchronous transmission and tdma,'' in \emph{DCOSS}.\hskip 1em plus
  0.5em minus 0.4em\relax IEEE, 2017, pp. 19--26.

\bibitem{almalkawi2010wireless}
I.~T. Almalkawi, M.~Guerrero~Zapata, J.~N. Al-Karaki, and J.~Morillo-Pozo,
  ``Wireless multimedia sensor networks: current trends and future
  directions,'' \emph{Sensors}, vol.~10, no.~7, pp. 6662--6717, 2010.

\bibitem{theNTP}
D.~L. {Mills}, ``Internet time synchronization: the network time protocol,''
  \emph{IEEE Transactions on Communications}, vol.~39, no.~10, 1991.

\end{thebibliography}
